%% file: sn-article.tex
\documentclass[pdflatex,iicol,sn-basic]{sn-jnl}% Style for submissions to Nature Portfolio journals
\usepackage[table]{xcolor}
\usepackage{graphicx}%
\usepackage{multirow}%
\usepackage{amsmath,amssymb,amsfonts}%
\usepackage{amsthm}%
\usepackage{mathrsfs}%
\usepackage[title]{appendix}%
\usepackage{xcolor}%
\usepackage{textcomp}%
\usepackage{manyfoot}%
\usepackage{booktabs}%
\usepackage{algorithm}%
\usepackage{algorithmicx}%
\usepackage{algpseudocode}%
\usepackage{multirow}
\usepackage{booktabs}
\usepackage{makecell}
\usepackage{listings}%
\theoremstyle{thmstyleone}%
\theoremstyle{thmstyletwo}%

\theoremstyle{thmstylethree}%

\begin{document}

\title[Article Title]{EvoTale: Continual Character Customization for Expanding Story Worlds}

\author[1]{\fnm{Jinlu} \sur{Zhang}}

\author[1]{\fnm{Qiyun} \sur{Wang}}

\author[1]{\fnm{Baoxiang} \sur{Du}}

\author[1]{\fnm{Jiayi} \sur{Ji}}

\author[2]{\fnm{Jing} \sur{He}}

\author[3]{\fnm{Rongsheng} \sur{Zhang}}

\author[3]{\fnm{Tangjie} \sur{Lv}}

\author*[1]{\fnm{Xiaoshuai} \sur{Sun}}

\author[1]{\fnm{Rongrong} \sur{Ji}}

\affil[1]{\orgdiv{Key Laboratory of Multimedia Trusted Perception and Efficient Computing}, \orgname{Ministry of Education of China, Xiamen
University}, \postcode{361005}, \country{P.R., China}}

\affil[2]{\orgname{The Hong Kong University
of Science and Technology (Guangzhou)}, \city{Guangzhou}, \postcode{510000}, \country{P.R., China}}

\affil[3]{\orgdiv{Fuxi AI Lab}, \orgname{NetEase Inc.}, \orgaddress{\city{Hangzhou}, \postcode{310000}, \country{P.R., China}}}
\input{sec/0_abs}
\maketitle

\input{sec/1_intro}

\input{sec/2_rela}
\input{sec/3_method}
\input{sec/4_exp}

\input{sec/5_con}

\bibliography{sn-bibliography}% common bib file
\end{document}

%% file: sec/0_abs.tex
\abstract{
Character-centric story visualization aims to synthesize coherent image sequences that depict narrative events and interactions while preserving recurring character identities. In expanding story worlds, new user-specified characters must be continually incorporated despite varying customization difficulty and identity conflicts in multi-character scenes, without disrupting previously learned identities. In this paper, we propose EvoTale, a continual character customization framework for expanding stylized story worlds. We first introduce an All-in-One-World Character Integrator, which accumulates character-specific residual components within a unified LoRA branch using sparsely overlapping subspaces spanned by a shared orthonormal basis, while freezing previously learned components to limit cross-character coupling. We then develop a Character Quality Gate that uses rubric-guided MLLM feedback as a bounded controller to adapt the optimization budget based on the assessed customization quality. Finally, we propose Character-Aware Region-Focus Sampling, which combines bounding-box-guided regional denoising with identity-aware global denoising to preserve character identities within their designated regions while maintaining global narrative coherence. Experimental results show that EvoTale achieves a favorable balance across character fidelity, continual identity retention, multi-character generation quality, and story-text alignment compared with representative story visualization and customization methods.
}

\keywords{Image Generation, Diffusion Model, Story Visualization, Continual Character Customization}

%% file: sec/1_intro.tex
\section{Introduction}

\begin{figure*}[!t]
    \centering
    \includegraphics[width=1.0\textwidth]{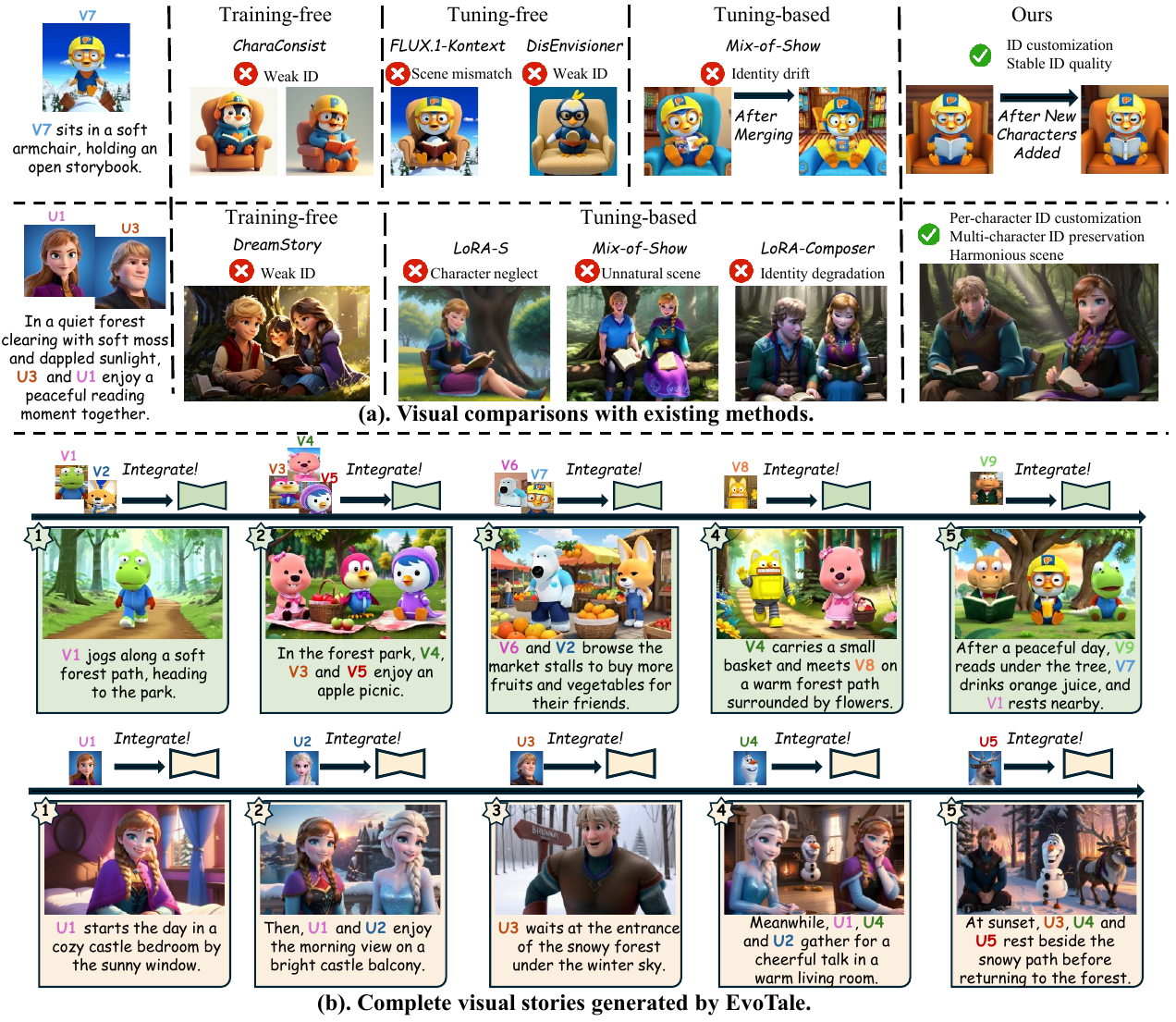}
    \caption{
Visual comparisons and continual character integration in expanding stylized story worlds.
\textbf{(a)} Compared with representative methods, EvoTale better preserves user-specified identities and scene coherence in single- and multi-character settings.
\textbf{(b)} EvoTale continually expands the character set while retaining previously integrated identities, as illustrated by \textcolor[RGB]{33,95,154}{V2} after adding \textcolor[RGB]{87,156,219}{V7} in \textit{Pororo} and \textcolor[RGB]{33,95,154}{U2} after adding \textcolor[RGB]{59,125,35}{U4} in \textit{Frozen}.
}
    \label{fig:fig1}
\end{figure*}

Story visualization is an important task in computer vision with broad application potential across creative media production, visual content authoring, and visual communication~\citep{li2019storygan,song2020character,tong2018storytelling,bugliarello2023storybench,huang2024story3d,ma2011scientific}. Given a multi-sentence narrative, story visualization aims to synthesize a sequence of story frames that faithfully depict the events, scenes, and character interactions described in the text while preserving the visual identities of recurring characters across frames~\citep{li2019storygan,zhou2024storydiffusion,zhang2025storyweaver}. A key challenge is therefore to jointly achieve accurate story-text alignment and cross-frame character consistency~\citep{zhang2025storyweaver,dong2025storycrafter,zhuang2025vistorybench,pan2024synthesizing}, which becomes more demanding when new user-specified characters are incorporated sequentially into existing narratives, since each new character must be integrated without compromising previously learned character identities~\citep{wang2024evolving,dong2024continually,guo2025conceptguard}.

Existing story visualization methods can be broadly categorized into two types. Open-ended methods~\citep{liu2024intelligent,he2025dreamstory,zhou2024storydiffusion,jeong2023zero,wang2025characonsist} generate visual narratives without a predefined cast, typically conditioning on narrative text, reference images, or internally generated cross-frame context. Though flexible across unseen stories and characters, they provide limited control over fine-grained user-specified characters. As a result, semantically plausible frames may deviate from the specified identities, as shown by the training-free results in Fig.~\ref{fig:fig1}(a). Closed-world methods~\citep{li2019storygan,song2020character,chen2022character,li2022word,zhang2025storyweaver,tao2024storyimager,shen2025storygpt} are trained or adapted for a predefined cast, enabling stable character representations within a specific story world. However, extending the cast after training generally requires additional model adaptation.

A natural alternative is to incorporate image customization methods~\citep{ruiz2023dreambooth,gal2022image,yang2024lora,smith2023continual,guo2025conceptguard,dong2024continually,labs2025flux} into story generation. However, existing approaches address only subsets of the requirements for expanding story worlds. Reference-conditioned methods~\citep{labs2025flux,ye2023ip,he2024disenvisioner} inject identities without per-character optimization, but may lose fine-grained details or remain overly tied to reference appearances under changes in pose, viewpoint, scene, and interaction. Multi-concept customization methods~\citep{gu2023mix,kumari2023multi,yang2024lora} support the composition of multiple customized concepts, but may suffer from identity drift, character neglect, or scene incoherence, as illustrated by the tuning-free and tuning-based results in Fig.~\ref{fig:fig1}(a). Continual customization methods~\citep{dong2024continually,guo2025conceptguard,smith2023continual} enable sequential concept integration while mitigating forgetting. Nevertheless, existing methods remain limited in handling varying customization difficulty across characters and maintaining narrative consistency in multi-character generation. A uniform adaptation budget may underfit difficult identities while overfitting simpler ones. A practical framework should therefore combine continual character integration, quality-aware adaptation, and coherent multi-character composition with reliable identity assignment and global narrative coherence.

To address these limitations, we propose EvoTale, a continual story character customization framework for expanding stylized story worlds. EvoTale continually integrates new user-specified characters while preserving previously learned identities and supporting their coherent composition in multi-character story frames. Specifically, we first develop an \textbf{All-in-One-World Character Integrator}, which maintains a unified LoRA branch and incrementally incorporates character-specific residual components into sparsely overlapping subspaces spanned by a shared orthonormal basis. Previously learned components remain frozen, limiting cross-character coupling to shared basis directions. This unified parameterization reduces interference and eliminates the need for post-hoc merging of independently trained adapters. To accommodate differences in character customization difficulty, we further introduce a \textbf{Character Quality Gate}, which uses rubric-guided multimodal large language model (MLLM) assessment solely as a bounded training controller to evaluate adaptation quality and adjust the training budget when necessary. Finally, for story frames involving multiple customized characters, we propose \textbf{Character-Aware Region-Focus Sampling}. It uses coarse bounding boxes to guide character-specific denoising within local regions, while global denoising captures the overall scene composition and interactions among characters. This sampling strategy improves character assignment and identity preservation while maintaining scene coherence and global narrative semantics (Fig.~\ref{fig:fig1}(b)).

We evaluate EvoTale under a continual character-learning protocol in the \textit{Pororo} and \textit{Frozen} story worlds in TBC-Bench~\citep{zhang2025storyweaver}, covering both single- and multi-character story visualization, and compare it against representative story visualization and concept customization methods
~\citep{hu2022lora,gu2023mix,yang2024lora,gal2022image,dong2024continually,zhong2024multi,he2024disenvisioner,labs2025flux,zhang2025storyweaver,po2024orthogonal,wang2025characonsist,he2025dreamstory}. In SD1.5, EvoTale achieves the highest single-character D-I scores in both story worlds, with gains of 4.71\% on Pororo and 3.77\% on Frozen. It also exhibits limited identity degradation as characters are integrated sequentially. For multi-character generation, EvoTale achieves a favorable balance across character alignment, semantic alignment, and visual storytelling quality.

In summary, the contributions of this paper are threefold:
\begin{itemize}
\item We propose EvoTale, a framework for continual story character customization in expanding stylized story worlds, enabling the integration of new user-specified characters while preserving previously integrated identities.

\item We develop an \textbf{All-in-One-World Character Integrator} with orthogonal-basis-guided residual accumulation and character-conditioned routing, a \textbf{Character Quality Gate} for quality-aware adaptation and \textbf{Character-Aware Region-Focus Sampling} for identity-preserving and globally coherent multi-character story generation.

\item Extensive experiments show that EvoTale achieves a favorable balance across character fidelity, story-text alignment and continual identity retention.
\end{itemize}

%% file: sec/2_rela.tex
\section{Related Work}
\subsection{Story Visualization}
Story visualization aims to synthesize coherent image sequences from multi-sentence narratives
\citep{li2019storygan,tao2024storyimager,shen2025storygpt,zhang2025storyweaver,zheng2025contextualstory}. Existing methods can be broadly grouped into closed-world and open-ended paradigms according to whether they assume a predefined character set. Closed-world methods~\citep{li2019storygan,tao2024storyimager,shen2025storygpt,zhang2025storyweaver,zheng2025contextualstory} are trained or adapted for a particular story world and do not natively support persistent expansion to arbitrary user-specified characters, and extending the cast generally requires additional adaptation or retraining. Open-ended methods~\citep{liu2024intelligent,he2025dreamstory,zhou2024storydiffusion,zhou2025agentstory} condition on narrative text, reference images, or cross-frame context to support generation across unseen characters. However, these methods mainly preserve character consistency within the current story generation process, without persistently integrating newly introduced character identities for reuse in future stories. In contrast, EvoTale targets continual character customization for stylized story visualization, allowing the cast to expand while retaining previously integrated identities.

\begin{figure*}[!t]
    \centering
    \includegraphics[width=1.0\textwidth]{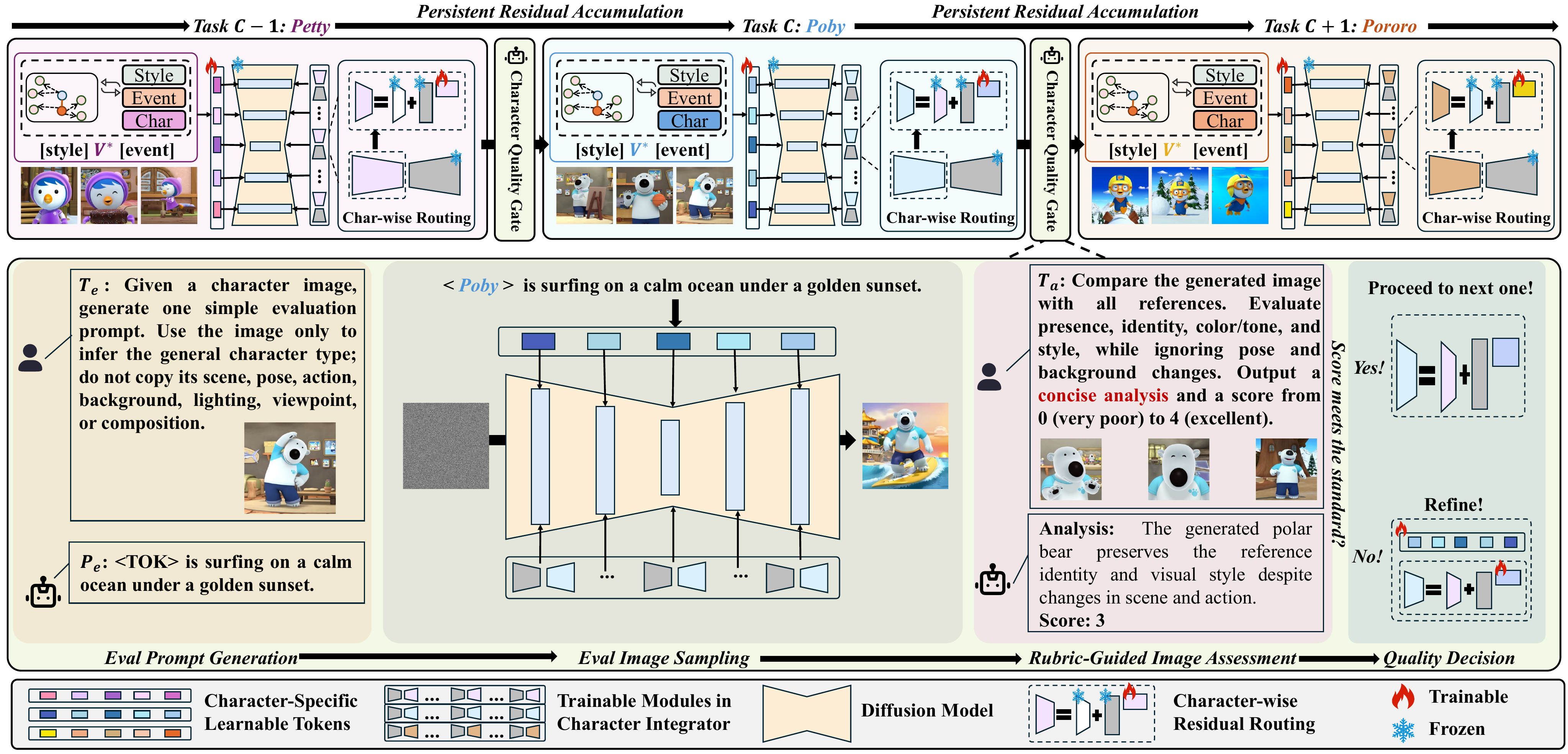}
\caption{
\textbf{Overview of EvoTale's continual character customization pipeline.}
The All-in-One-World Character Integrator sequentially incorporates newly introduced characters into a unified framework while preserving previously acquired character knowledge. After each customization attempt, the Character Quality Gate (CQG) employs an MLLM as a quality judge to generate an evaluation prompt, assess the customized result, and determine whether to proceed to the next character or further refine the current one.
}
    \label{fig:method1-overview}
\end{figure*}
\subsection{Image Customization}
Image customization aims to adapt pretrained text-to-image models to faithfully synthesize user-specified visual concepts from only a few reference images~\citep{ruiz2023dreambooth,gal2022image}. Tuning-free methods~\citep{wei2023elite,he2024disenvisioner,wang2024instantid,xiao2025fastcomposer,ye2023ip,labs2025flux} condition generation on reference images without subject-specific test-time optimization. Although efficient, they may either lose fine-grained identity details or inherit reference-specific appearance cues, limiting flexibility under substantial changes in pose, viewpoint, scene, or interaction. Tuning-based methods instead optimize subject- or domain-specific representations or parameters from limited reference data~\citep{ruiz2023dreambooth,gal2022image,2025zhuDomainStudio}, allowing fine-grained visual characteristics to be encoded through optimization. While effective for individual concepts, extending tuning-based customization to multiple concepts often requires joint or repeated optimization, or additional model-composition mechanisms~\citep{gu2023mix,kumari2023multi}.

\paragraph{Multi-Concept Customization.}
Multi-concept customization extends personalized image generation to jointly synthesize multiple user-specified concepts. Existing methods rely on joint optimization~\citep{kumari2023multi,han2023svdiff}, post-hoc model merging~\citep{gu2023mix,po2024orthogonal,hoang2025showflow}, or inference-time composition~\citep{kwon2024concept,kong2024omg,yang2024lora}. Orthogonal Adaptation~\citep{po2024orthogonal}, for example, encourages independently learned weight residuals to be mutually orthogonal, reducing interference during model aggregation. However, such approaches still rely on independently trained concept-specific components and require an explicit aggregation or composition step. When customized concepts arrive sequentially, continual learning (CL) provides a natural paradigm for learning new knowledge while preserving previously learned capabilities~\citep{wang2024comprehensive,cha2021co2l,wang2022learning,kong2023trust,cotogni2025exemplar}. Recent continual customization methods extend the CL paradigm to support sequential concept learning while mitigating interference with previously acquired concepts~\citep{dong2024continually,guo2025conceptguard,smith2023continual}. However, existing methods mainly focus on retaining previously learned concepts during continual customization, rather than adapting the learning process to the needs of each newly introduced concept. In contrast, EvoTale first integrates new characters within a single persistent LoRA branch. The character-specific residual components are assigned to sparsely overlapping subspaces spanned by a shared orthonormal basis, while previously learned components remain frozen, limiting cross-character coupling to shared basis directions. EvoTale further adjusts the customization budget according to the assessed quality of each character.

Beyond maintaining multiple customized concepts in a unified model, another challenge is to compose multiple customized characters within a single image, where concept neglect, identity mixing, and cross-concept interference often arise~\citep{xiao2025fastcomposer,gu2023mix,yang2024lora,jiang2025mc}. Existing methods address these issues through inference-time composition or optimization~\citep{zhong2024multi,jin2025latexblend,jiang2025mc}, attention-based identity control~\citep{he2025dreamstory,zhao2026freercustom,banerjee2025talediffusion}, and explicit spatial guidance such as masks, key-poses, or bounding boxes~\citep{zhang2025storyweaver,dong2024continually,gu2023mix,banerjee2025talediffusion}. These mechanisms are often combined to improve concept separation and identity assignment. Story visualization further requires multi-character composition to reflect narrative context and inter-character relations~\citep{he2025dreamstory,zhang2025storyweaver,zhou2025agentstory,banerjee2025talediffusion}. EvoTale uses coarse bounding boxes to guide character-specific regional denoising, while global denoising captures the overall scene structure and inter-character interactions. This design improves identity assignment and preservation while maintaining scene coherence.

%% file: sec/3_method.tex
\section{Method}
\subsection{Overview}
EvoTale adopts the Character-Graph from StoryWeaver~\citep{zhang2025storyweaver}, with $G=\langle O,E,A\rangle$, where $O$, $E$, and $A$ denote characters, events, and associated attributes. We retain this semantic representation and focus on continual character customization for an expanding character cast. Unlike StoryWeaver, whose generator is optimized for a predefined character set, EvoTale sequentially acquires new character identities while preserving previously learned ones.

As illustrated in Fig.~\ref{fig:method1-overview} and Fig.~\ref{fig:multi-sampling}, EvoTale consists of three complementary components. First, the All-in-One-World Character Integrator (\textbf{CI}) accumulates character-specific residuals in a persistent LoRA branch and uses character-conditioned subspace routing during customization and inference. Second, the Character Quality Gate (\textbf{CQG}) uses an MLLM-based quality signal to accept the current customization or trigger bounded adaptive refinement. Finally, Character-Aware Region-Focus (\textbf{CA-RF}) Sampling combines Role-Specific Crop Denoising (\textbf{RS-CD}) for identity anchoring with Identity-Aware Global Denoising (\textbf{IA-GD}) for global harmonization through time-aware fusion.

\subsection{All-in-One-World Character Integrator}
\label{sec:ci}

Following prior methods~\citep{gu2023mix,po2024orthogonal,dong2024continually}, we adopt LoRA~\citep{hu2022lora} to encode character-specific visual knowledge. For a linear layer with weight $W$, LoRA introduces a low-rank residual $\Delta W=AB$, where $A\in\mathbb{R}^{d\times r}$ and $B\in\mathbb{R}^{r\times k}$ denote the down- and up-projection matrices, respectively. Unlike conventional subject customization, continual story character customization must incorporate new characters as the cast expands while preserving previously integrated identities for reuse in subsequent story frames. This motivates a residual parameterization that supports continual accumulation with limited cross-character interference.

Existing multi-concept methods typically learn separate concept-specific modules and combine them when needed~\citep{po2024orthogonal,hoang2025showflow,gu2023mix}, whereas continual customization methods sequentially incorporate new concepts while preserving previously learned ones~\citep{dong2024continually}. In particular, Orthogonal Adaptation~\citep{po2024orthogonal} trains separate concept-specific residuals using bases sampled from a shared orthogonal matrix and aggregates the resulting modules after customization. Building on its randomized orthogonal-basis principle, our All-in-One-World Character Integrator (\textbf{CI}) instead organizes all character-specific increments within a unified LoRA branch.

Specifically, for each LoRA-injected linear layer, we construct a fixed orthonormal basis $Q\in\mathbb{R}^{d\times d}$ by applying QR decomposition to a matrix whose entries are independently sampled from a standard Gaussian distribution, such that $Q^{\top}Q=I_d$. For each incoming character $c_i$, CI randomly and uniformly samples $r_1$ distinct column indices to form an index set $\mathcal{S}_i$ and defines the corresponding character-specific basis as:
\begin{equation}
D_i=Q[:,\mathcal{S}_i],
\qquad
P_i=D_iD_i^{\top},
\qquad
|\mathcal{S}_i|=r_1,
\label{eq:ci_basis_selection}
\end{equation}
where $D_i^{\top}D_i=I_{r_1}$, and $P_i$ denotes the projection matrix associated with $c_i$. The shared basis $Q$ remains fixed throughout continual customization, while each $D_i$ selects a character-specific subset of its basis directions. For two independently sampled subsets $S_i$ and $S_j$, the expected overlap is
$\mathbb{E}[|S_i\cap S_j|]=r_1^2/d$, corresponding to a fraction of $r_1/d$ relative to each character subspace. Therefore, when $r_1\ll d$, the character subspaces share only a small fraction of their assigned directions.

CI accumulates character-specific knowledge in the down-projection factor $A$, while the shared up-projection factor $B$ remains fixed. After integrating $i$ characters, the accumulated factor is:
\begin{equation}
A^{(i)}
=
\sum_{k=1}^{i}D_kL_k,
\qquad
\Delta W^{(i)}=A^{(i)}B,
\label{eq:ci_accumulation}
\end{equation}
where $L_k\in\mathbb{R}^{r_1\times r}$ denotes the learnable coefficient matrix for character $c_k$. Each character contributes an additive component $D_kL_k$ to the unified LoRA branch. When a new character $c_i$ is incorporated, its component $D_iL_i$ is added to $A^{(i-1)}$, while all previously learned components remain fixed. The shared matrix $B$ is initialized once with entries sampled from a standard Gaussian distribution and remains fixed throughout continual customization.

Following ~\citep{gu2023mix,dong2024continually}, each character $c_j$ is also associated with a set of layer-wise learnable character embeddings $\{v_j^{\ell}\}_{\ell=1}^{L}$. When customizing $c_j$, $L_j$ and its character embeddings are jointly optimized, while the shared matrix $B$, all previous coefficients $\{L_k\}_{k<j}$, and previously learned character embeddings remain fixed. CI then routes the accumulated LoRA residual through the character-specific projector $P_j$, yielding:
\begin{equation}
XW
+
XP_j
\left(
A^{(j-1)}+D_jL_j
\right)B.
\label{eq:ci_training_forward}
\end{equation}
Using $P_j=D_jD_j^{\top}$, $D_j^{\top}D_j=I_{r_1}$, and
$A^{(j-1)}=\sum_{k<j}D_kL_k$, the routed residual is formulated as:
\begin{equation}
\begin{aligned}
&XP_j
\left(
A^{(j-1)}+D_jL_j
\right)B\\
&=
XD_j
\left(
D_j^{\top}A^{(j-1)}+L_j
\right)B \\
&=
XD_j
\left(
L_j+
\sum_{k<j}D_j^{\top}D_kL_k
\right)B.
\end{aligned}
\label{eq:ci_training_routing}
\end{equation}
In Equ.\ref{eq:ci_training_routing}, the first term captures the character-specific residual learned for $c_j$, while the second term represents residual contributions from previously integrated characters that are routed into the subspace of $c_j$. Since the shared matrix $B$ and all previous coefficients $\{L_k\}_{k<j}$ remain fixed, optimizing $L_j$ cannot directly overwrite their stored residual components. Moreover, $D_j^{\top}D_k$ has nonzero entries only for basis directions shared by the two characters. Contributions from previous characters are therefore restricted to the sparse overlap between their assigned subspaces, allowing the new character to be learned primarily through its own basis directions.

After customizing $c_j$, its learned component is added to the unified LoRA branch:
\begin{equation}
A^{(j)}
=
A^{(j-1)}+D_jL_j
=
\sum_{k=1}^{j}D_kL_k.
\label{eq:ci_incremental_update}
\end{equation}
Both $L_j$ and its character embeddings are then fixed during all subsequent customization steps.

We next examine the reverse direction, namely the influence of a newly integrated character on previously learned characters. After incorporating $c_j$, the residual routed for an earlier character $c_i$, where $i<j$, is formulated as:
\begin{equation}
\begin{aligned}
&XP_iA^{(j)}B\\
&=
XP_iA^{(j-1)}B
+
XP_iD_jL_jB \\
&=
XP_iA^{(j-1)}B
+
XD_i
\left(
D_i^{\top}D_j
\right)
L_jB.
\end{aligned}
\label{eq:ci_backward_interference}
\end{equation}
The first term preserves the residual routed to $c_i$ before learning $c_j$, whereas the second represents the additional influence introduced by the new character. This additional term is restricted by $D_i^{\top}D_j$ to the basis directions jointly selected by the two characters. CI therefore combines parameter freezing with sparse, basis-dependent coupling to limit bidirectional cross-character interference, consistent with the limited identity degradation, as shown in Sec.~\ref{sec:exp_cl}.

After integrating $T$ characters, generating a target character $c_i$ uses the same character-specific subspace to route the unified LoRA residual:
\begin{equation}
XW
+
(XD_i)
\left(
D_i^{\top}A^{(T)}
\right)B.
\label{eq:ci_forward}
\end{equation}
This factorized form is equivalent to $XW+XP_iA^{(T)}B$, but avoids explicitly constructing the $d\times d$ projection matrix $P_i$. The routing is confined to the LoRA residual branch and does not modify the pretrained weight $W$.

In our implementation, CI is applied to the query, key, and value projections of self-attention layers and to the query projection of cross-attention layers. For clarity, we omit layer indices throughout the formulation.

\subsection{Character Quality Gate via MLLM-as-Judge}
\label{sec:evaluator}

Another challenge in continual character customization is determining whether the current character has been sufficiently customized before moving on. A fixed training schedule provides no explicit quality signal for this decision. Moreover, characters with different shapes, accessories, and visual styles may require different amounts of optimization. Consequently, a uniform training configuration may under-customize difficult identities while over-optimizing simpler ones. To address this issue, we introduce a Character Quality Gate (\textbf{CQG}), a bounded quality-aware controller that evaluates each customization attempt and determines whether to accept the current character or trigger further refinement with adjusted learning rates and training steps.

At the end of each customization attempt for character $c_i$, EvoTale randomly samples $n$ reference images $\mathcal{I}_i=\{I_i^j\}_{j=1}^{n}$ from the training images of $c_i$ in each of the evaluation trials. One image from $\mathcal{I}_i$, together with a prompt-generation instruction $T_e$, is provided to an MLLM~\citep{Qwen3-VL} to generate an evaluation prompt $P_e$ for subsequent image generation. The instruction asks the MLLM to construct a prompt that places the character in a new scene, action, or environment. The customized model then generates an evaluation image $I_e$ conditioned on $P_e$.

Inspired by DreamBench++~\citep{peng2024dreambench}, we also use the MLLM as a quality judge for character customization. The MLLM compares $I_e$ with all images in $\mathcal{I}_i$ under a rubric-guided instruction $T_a$ and assigns an integer score $s\in[0,4]$, corresponding to five quality levels from \emph{very poor} to \emph{excellent}. Our character-focused rubric considers four aspects, \emph{i.e.,} \emph{character presence}, \emph{character identity}, \emph{color and tone consistency}, and \emph{style representation}. The score serves only as an internal control signal for CQG and is neither treated as an absolute measure of customization quality nor used as a final evaluation metric.

\begin{figure*}[!t]
    \centering
    \includegraphics[width=1.0\textwidth]{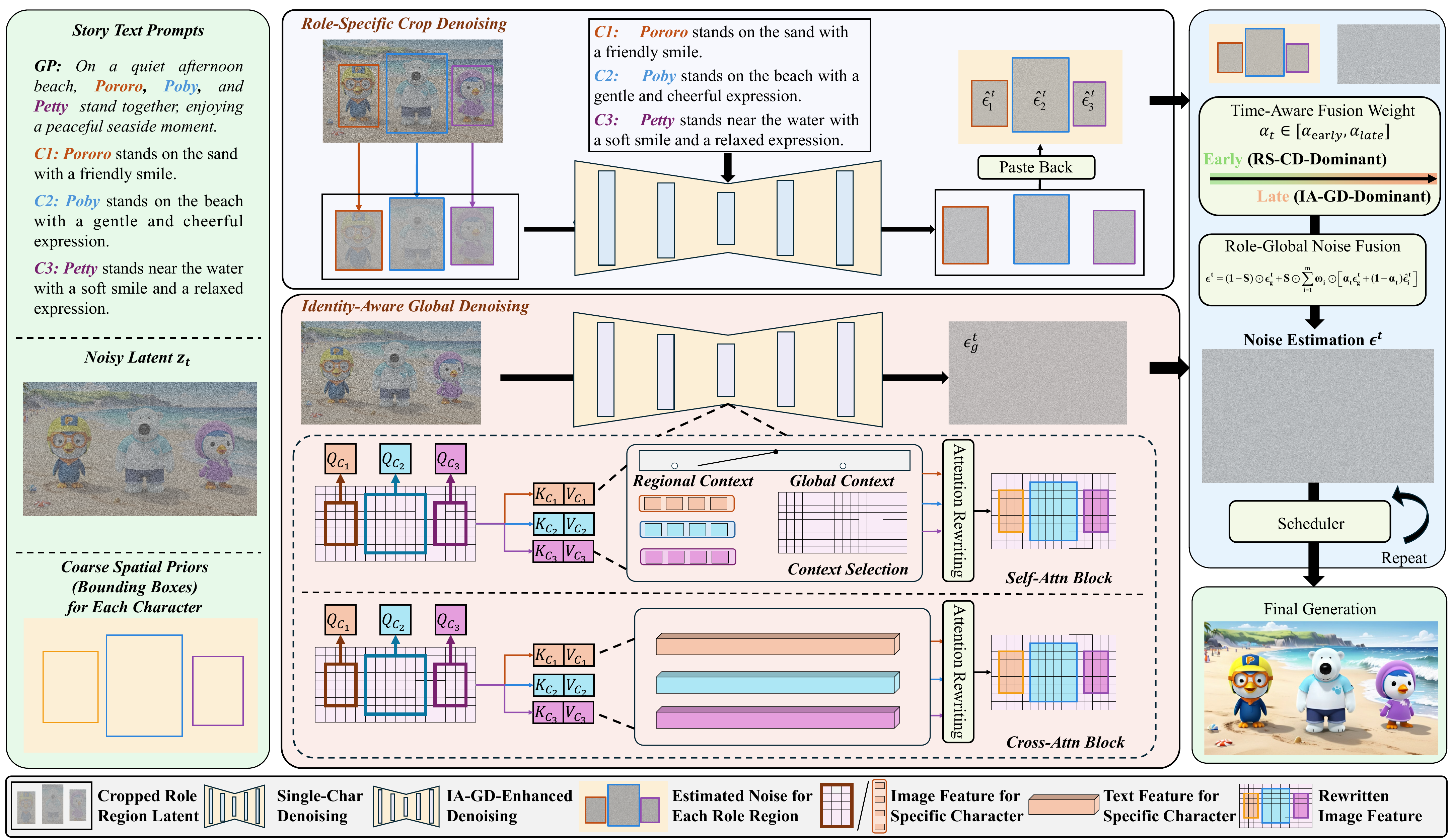}
\caption{
\textbf{Overview of Character-Aware Region-Focus Sampling.}
RS-CD predicts character-specific regional noise, while IA-GD preserves global context through character-conditioned attention. A time-aware schedule initially favors RS-CD for identity preservation and gradually shifts toward IA-GD for coherent scene composition.
}
    \label{fig:multi-sampling}
\end{figure*}

To stabilize this quality signal, CQG repeats the complete evaluation procedure $K$ times and computes the mean score for the $m$-th customization attempt as $\bar{s}^{(m)}
=
\frac{1}{K}
\sum_{\ell=1}^{K}s_{\ell}^{(m)}.$
The mean score is compared against a predefined threshold $\tau$. If $\bar{s}^{(m)}\geq\tau$, the current customization is accepted, and training proceeds to the next character. Otherwise, CQG triggers another refinement attempt and computes a score-adaptive refinement factor:
\begin{equation}
\rho^{(m)}
=
\max\left(
\rho_{\min},
1-\exp\left(\bar{s}^{(m)}-\tau\right)
\right).
\end{equation}
The learning rate and number of training steps for the next attempt are updated as:
\begin{equation}
\left\{
\begin{aligned}
lr^{(m+1)}
&=
\max\left(
lr_{\min},
\rho^{(m)}lr^{(m)}
\right),\\
step^{(m+1)}
&=
\max\left(
step_{\min},
\rho^{(m)}step^{(m)}
\right).
\end{aligned}
\right.
\label{equ:modify_cfg}
\end{equation}
Here, $lr_{\min}$ and $step_{\min}$ are derived from the initial customization configuration using the same lower-bound factor $\rho_{\min}$, rather than introduced as additional tuning parameters. Since the refinement factor is applied recursively across successive attempts, these bounds prevent the optimization budget from vanishing after repeated scaling. A lower quality score produces a larger $\rho^{(m)}$, retaining more of the current optimization scale for further customization, whereas a score closer to $\tau$ results in a lighter refinement. This yields a bounded coarse-to-fine schedule in which refinement becomes progressively more conservative without degenerating into negligible updates.

Finally, CQG allows at most $M$ customization attempts. The threshold $\tau$ therefore serves as an early-acceptance target rather than a quality guarantee. The current customization is accepted once $\bar{s}^{(m)}\geq\tau$. Otherwise, if the threshold is not reached within $M$ attempts, CQG retains the latest state and proceeds to the next character. This bounded design limits training overhead while allowing underperforming characters to receive controlled refinement.

\subsection{Character-Aware Region-Focus Sampling}
\label{sec:sampling}

Generating multiple customized characters in a single story frame requires simultaneously preserving individual identities and maintaining a coherent global composition. A straightforward strategy is to combine role-specific crop predictions with standard full-image denoising using a fixed fusion weight. Although this provides strong local identity guidance, it may lead to rigid character layouts. Replacing the fixed weight with a time-aware schedule improves global composition, but the increasing contribution of standard global denoising may weaken identity-specific guidance because it lacks character-aware routing. These two strategies are illustrated by the first two settings in Fig.~\ref{fig:abla}(b).

To balance these objectives, we introduce Character-Aware Region-Focus (\textbf{CA-RF}) Sampling, which performs two complementary denoising processes in parallel from the same noisy latent as shown in Fig.\ref{fig:multi-sampling}. A \emph{Role-Specific Crop Denoising} (\textbf{RS-CD}) branch provides strong local identity cues through single-character crop denoising, while an \emph{Identity-Aware Global Denoising} (\textbf{IA-GD}) branch performs character-aware full-image denoising through identity-conditioned attention routing. Their predictions are fused with a time-aware schedule, progressively shifting from RS-CD-dominant identity anchoring to IA-GD-dominant global harmonization.

For a frame containing $m$ characters, CA-RF takes
$\{\mathcal{P}_g,\{\mathcal{P}_i\}_{i=1}^{m},\{\mathcal{B}_i\}_{i=1}^{m}\}$ as input, where $\mathcal{P}_g$ denotes the global story prompt, $\mathcal{P}_i$ is the character-specific prompt for $c_i$, and $\mathcal{B}_i=(h_i^s,w_i^s,h_i^e,w_i^e)$ denotes its bounding box. The global prompt provides the overall story context, while each character-specific prompt provides the identity condition for its corresponding character. The bounding boxes serve as coarse spatial priors that associate these character-specific conditions with their designated regions.

\noindent \textbf{Role-Specific Crop Denoising (RS-CD).}
Inspired by the region noise estimation strategy in CIDM~\citep{dong2024continually}, RS-CD adopts a different crop-wise formulation that explicitly isolates each character during noise prediction. Rather than estimating region-conditioned noise on the full latent, RS-CD crops each character region from the current noisy latent and performs an independent single-character denoising forward.

For any tensor $U$ with spatial resolution $H_U\times W_U$, we rescale the normalized bounding box $\mathcal{B}_i$ to the resolution of $U$ and denote the result by:
\begin{equation}
\mathcal{B}_i^{U}
=
\operatorname{Scale}(\mathcal{B}_i,H_U,W_U).
\end{equation}
This notation applies to both latent-level cropping and feature-level regional processing. 

Given the noisy latent $z_t$, RS-CD extracts the latent region associated with character $c_i$ as:
\begin{equation}
z_{t,i}
=
\operatorname{Crop}
(z_t,\mathcal{B}_i^{z_t}),
\end{equation}
and independently predicts the role-specific noise using the corresponding single-character condition in CI routing:
\begin{equation}
\epsilon_i^t
=
\epsilon_{\theta}
(z_{t,i},t,\mathcal{P}_i,c_i).
\end{equation}
The predicted crop noise is then mapped back to its original location on a full-size noise canvas, with zeros elsewhere:
\begin{equation}
\hat{\epsilon}_i^t
=
\operatorname{Paste}
(\epsilon_i^t,\mathcal{B}_i^{z_t}).
\end{equation}
This procedure constructs a role-specific noise prediction, providing a strong character-specific signal for identity anchoring.

\noindent \textbf{Identity-Aware Global Denoising (IA-GD).}
To complement the isolated local predictions of RS-CD with full-image context, IA-GD first computes global attention and then enhances character regions using identity-conditioned attention. Let $X$ denote an intermediate image feature. The standard attention mechanism is formulated as:
\begin{equation}
\operatorname{Attn}(Q,K,V)
=
\sigma\left(
\frac{QK^{T}}{\sqrt{d}}
\right)V
\end{equation}
In this case, the global self- and cross-attention outputs are formulated as:
\begin{equation}
\left\{
\begin{aligned}
O_g^s
&=
\operatorname{Attn}
(XW_q^s,XW_k^s,XW_v^s),\\
O_g^c
&=
\operatorname{Attn}
(XW_q^c,f_gW_k^c,f_gW_v^c),
\end{aligned}
\right.
\label{equ:global_attn}
\end{equation}
where $f_g$ denotes the global text feature.

For each character $c_i$, we extract its region at the current feature resolution:
\begin{equation}
X_i
=
\operatorname{Crop}
(X,\mathcal{B}_i^{X}).
\end{equation}
Let $\Phi_{q,i}^{s}$, $\Phi_{k,i}^{s}$, and $\Phi_{v,i}^{s}$ denote the CI-enhanced self-attention query, key, and value projections for character $c_i$, respectively, and let $\Phi_{q,i}^{c}$ denote its enhanced cross-attention query projection:
\begin{equation}
\begin{aligned}
Q_i^s &= \Phi_{q,i}^{s}(X_i),
&
Q_i^c &= \Phi_{q,i}^{c}(X_i),\\
K_{i,r}^s &= \Phi_{k,i}^{s}(X_i),
&
V_{i,r}^s &= \Phi_{v,i}^{s}(X_i),\\
K_{i,g}^s &= \Phi_{k,i}^{s}(X),
&
V_{i,g}^s &= \Phi_{v,i}^{s}(X),
\end{aligned}
\end{equation}
where $\Phi_{\cdot,i}$ follows the character-conditioned routing introduced in Sec.~\ref{sec:ci}.
For self-attention, IA-GD adopts a local-to-global schedule as:
\begin{equation}
O_i^s =
\begin{cases}
\operatorname{Attn}(Q_i^s,K_{i,r}^s,V_{i,r}^s),
& t>T_{\theta},\\
\operatorname{Attn}(Q_i^s,K_{i,g}^s,V_{i,g}^s),
& t\leq T_{\theta},
\end{cases}
\label{equ:iagd_self}
\end{equation}
where $T_{\theta}$ controls the transition from regional to global key-value context. During early denoising, restricting the keys and values to the character region strengthens character-specific feature formation. As denoising proceeds, expanding the key-value context to the full feature map allows each character to interact with the surrounding scene, while the region-specific query preserves its identity focus. This transition facilitates more natural character placement and scene composition as shown in Fig.\ref{fig:abla}(b).

For cross-attention, $Q_i^c$ attends to the character-specific text feature $f_i$, yielding:
\begin{equation}
O_i^c
=
\operatorname{Attn}
\left(
Q_i^c,
f_iW_k^c,
f_iW_v^c
\right).
\label{equ:iagd_cross}
\end{equation}

The character-specific outputs $\{O_i^s,O_i^c\}_{i=1}^{m}$ replace the corresponding regions in $\{O_g^s,O_g^c\}$, while the global attention responses are retained elsewhere. The resulting attention features are propagated through the full-image denoising forward to produce the full-image noise prediction $\epsilon_g^t$. IA-GD thus preserves full-image contextual modeling while maintaining identity-aware processing within character regions.

\noindent\textbf{Time-Aware Role-Global Fusion.}
To balance character identity preservation and global scene coherence, CA-RF dynamically fuses the character-specific noise predictions from RS-CD with the global character-enhanced noise prediction from IA-GD throughout denoising. Let $M_i$ denote the feathered soft mask derived from $B_i$ on the latent canvas. We define:
\begin{equation}
\begin{aligned}
S &=
\operatorname{clip}\left(
\sum_{i=1}^{m} M_i,\,0,\,1
\right),\\
\omega_i &=
\frac{M_i}{\sum_{j=1}^{m} M_j},
\qquad S>0.
\end{aligned}
\label{eq:spatial_weight}
\end{equation}
where $S$ denotes the aggregate character-region mask and $\omega_i$ denotes the normalized spatial contribution of character $c_i$. Outside the character regions ($S=0$), we simply set $\omega_i=0$. To progressively shift the fusion from character-specific denoising to global denoising, we define the time-aware fusion scale as:
\begin{equation}
\alpha_t
=
\alpha_{\mathrm{early}}
+
\left(
\alpha_{\mathrm{late}}
-
\alpha_{\mathrm{early}}
\right)
\pi_t^{\gamma},
\end{equation}
where $T$ denotes the maximum diffusion timestep and
$\pi_t=(T-t)/T\in[0,1]$ denotes the normalized denoising progress, which increases as denoising proceeds. The exponent $\gamma$ controls the transition rate, and we set
$0\leq\alpha_{\mathrm{early}}<\alpha_{\mathrm{late}}\leq1$.
Accordingly, a smaller $\alpha_t$ emphasizes the role-specific prediction $\hat{\epsilon}_i^t$ during early denoising, whereas its gradual increase shifts the fusion toward the global prediction $\epsilon_g^t$ at later steps.

The fused noise prediction is formulated as:
\begin{equation}
\epsilon^t
=
(1-S)\odot\epsilon_g^t
+
S\odot
\sum_{i=1}^{m}
\omega_i\odot
\left[
\alpha_t\epsilon_g^t
+
(1-\alpha_t)\hat{\epsilon}_i^t
\right].
\label{equ:ca_rf_fusion}
\end{equation}
Outside the character regions,  $\epsilon_g^t$ is retained unchanged. Within these regions, $\omega_i$ determines the spatial contribution of each role-specific prediction, while $\alpha_t$ controls the time-dependent balance between RS-CD and IA-GD. The fused prediction $\epsilon^t$ is then used to update $z_t$.

%% file: sec/4_exp.tex
\section{Experiments}
\subsection{Experimental Settings}
\subsubsection{Implementation Details.}
We implement EvoTale under a continual character-learning protocol and conduct experiments independently in two story worlds, \emph{i.e.,} 9 characters in \textit{Pororo} and 5 characters in \textit{Frozen} from TBC-Bench\citep{zhang2025storyweaver}. We use SD1.5 with the Rev-Animated checkpoint\footnote{\url{https://civitai.com/models/7371/rev-animated}} as the default backbone. We additionally report EvoTale with SDXL~\citep{podell2023sdxl} using the Starlight checkpoint\footnote{\url{https://huggingface.co/stablediffusionapi/starlight-xl-v3/tree/main}}. Methods~\citep{labs2025flux,wang2025characonsist,he2025dreamstory} based on SDXL or DiT~\citep{Peebles_2023_ICCV} are included as cross-backbone references.

For each character $c_j$, we optimize $L_j$ and the layer-wise character tokens by minimizing the standard diffusion denoising MSE loss. Optimization is performed using AdamW with an initial learning rate of $5\times10^{-4}$.
 For the initial customization attempt, each character-specific training set is repeated 500 times, so that the initial training budget scales with the number of reference images. We set $r=80$ and $r_1=20$. For \textbf{CQG}, we use Qwen3-VL-4B-Instruct as the MLLM and set $K=5$, $\tau=3$ on a $0$--$4$ scale, $M=8$, $n=3$ independently sampled reference images per complete evaluation, and $\rho_{\min}=0.2$. For \textbf{CA-RF}, we set $T_{\theta}=780$ and use staged global prompts, specifying the scene early and introducing character actions and interactions later. We use $(\alpha_{\mathrm{early}},\alpha_{\mathrm{late}},\gamma)=(0.2,0.8,2.0)$ for SD1.5 and $(0.2,0.6,2.0)$ for SDXL. For methods requiring spatial priors, we follow their official default configurations. For EvoTale, coarse character bounding boxes are arranged approximately evenly from left to right according to prompt order and remain fixed during sampling. All experiments are conducted on 4 GeForce RTX 4090 GPUs.

\subsubsection{Comparison Methods.}
We benchmark EvoTale against four categories of comparison methods for single- and multi-character story visualization. These include tuning-free customization methods~\citep{labs2025flux,he2024disenvisioner}, tuning-based customization and composition methods~\citep{gu2023mix,yang2024lora,zhong2024multi,po2024orthogonal}, sequential or continual customization methods~\citep{gal2022image,dong2024continually}, and different story visualization methods~\citep{zhang2025storyweaver,he2025dreamstory,wang2025characonsist}. We categorize Textual Inversion~\citep{gal2022image} as a continual method because character-specific token embeddings can be learned sequentially without modifying previously acquired embeddings. Tuning-free methods are evaluated using their official implementations. For optimization-based baselines implemented in our experiments, we use the same character images and base model checkpoint as EvoTale.
\begin{table*}[!t]
    \centering
    \small
    \setlength{\tabcolsep}{4pt}
    \caption{
    Quantitative comparisons on single-character story settings.
    The upper block reports comparisons among SD1.5-based methods,
    while the lower block provides cross-backbone reference results.
    Best results are shown in \textbf{bold} and the second-best are \underline{underlined}.
    $\star$ denotes DiT-based models and $\dagger$ denotes SDXL-based models;
    unmarked methods use SD1.5.
    }
    
    \begin{tabular*}{\textwidth}{
        @{\extracolsep{\fill}}
        llcccccc
        @{}
    }
    \toprule
        \textbf{Type}
        & \textbf{Method}
        & \multicolumn{3}{c}{\textbf{Pororo}}
        & \multicolumn{3}{c}{\textbf{Frozen}} \\
    \cmidrule(lr){3-5}
    \cmidrule(lr){6-8}
        &
        & D-I$\uparrow$
        & C-I$\uparrow$
        & C-T$\uparrow$
        & D-I$\uparrow$
        & C-I$\uparrow$
        & C-T$\uparrow$ \\

    \midrule
        \multicolumn{8}{c}{\textit{SD1.5-Based Comparison}} \\
    \midrule

        Tuning-free
        & DisEnvisioner
        & 47.03
        & 74.76
        & 34.05
        & \underline{49.63}
        & \textbf{83.53}
        & 32.80 \\

        Tuning-based
        & Mix-of-Show
        & 49.89
        & 79.38
        & 34.17
        & 45.05
        & 80.66
        & 35.08 \\

        Tuning-based
        & Orth-Adapt
        & \underline{52.01}
        & \underline{79.99}
        & 34.06
        & 49.61
        & 82.56
        & 35.10 \\

        CL
        & Textual Inversion
        & 42.99
        & 73.32
        & \textbf{34.92}
        & 36.40
        & 73.40
        & \textbf{35.46} \\

        CL
        & CIDM
        & 47.91
        & 78.67
        & 32.65
        & 44.06
        & 80.85
        & 32.59 \\

        CL
        & Ours
        & \textbf{54.46}
        & \textbf{80.89}
        & \underline{34.42}
        & \textbf{51.50}
        & \underline{82.84}
        & \underline{35.23} \\

    \midrule
\multicolumn{8}{c}{\textit{Cross-Backbone References}} \\
    \midrule

        Story visualization(Training-free)
        & CharaConsist$^\star$
        & 49.52
        & 74.70
        & 34.68
        & 37.92
        & \underline{80.76}
        & 34.88 \\

        Tuning-free
        & FLUX.1 Kontext$^\star$
        & \textbf{55.06}
        & \underline{79.03}
        & \textbf{35.19}
        & \underline{45.08}
        & 79.88
        & \textbf{36.24} \\

        CL
        & Ours$^\dagger$
        & \underline{54.67}
        & \textbf{82.23}
        & \underline{35.04}
        & \textbf{52.78}
        & \textbf{85.67}
        & \underline{35.64} \\

    \bottomrule
    \end{tabular*}
    \label{tab:sin_quantitative}
\end{table*}
\begin{figure*}[!t]
    \centering
    \includegraphics[width=1.0\textwidth]{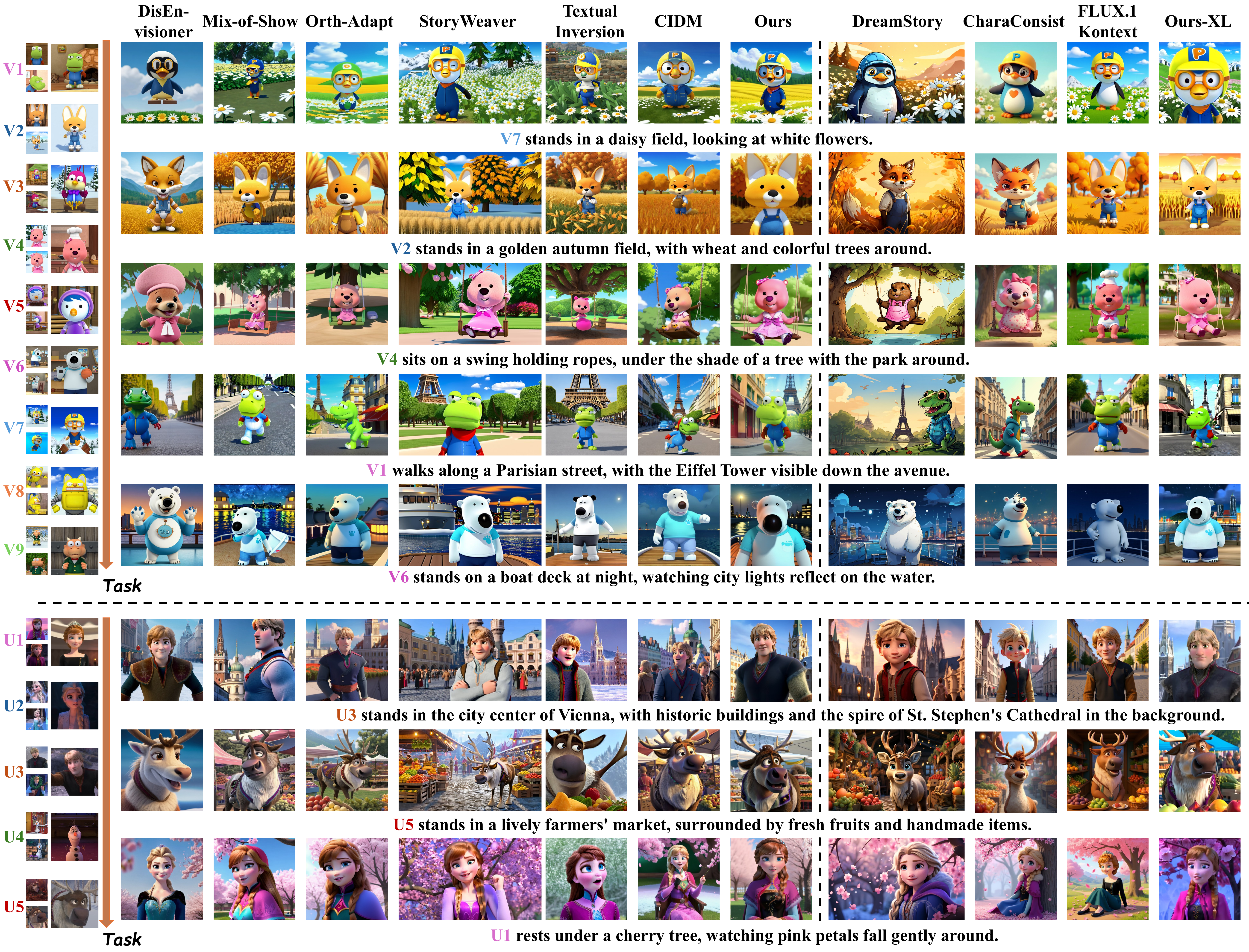}
\caption{
\textbf{Qualitative comparison of single-character story visualization.}
The arrows indicate the character learning order for continual-learning methods. EvoTale preserves character identity while maintaining favorable alignment with the story descriptions across both story settings.
}
    \label{fig:exp1_single}
\end{figure*}
\subsubsection{Evaluation Metrics.}
We evaluate character alignment and text--image semantic alignment for both settings. Character alignment is measured using full-frame DINO similarity (\textbf{D-I})~\citep{oquab2023dinov2} and CLIP image similarity (\textbf{C-I})~\citep{radford2021learning} between each generated frame and the  corresponding character reference images, while semantic alignment is measured using CLIP text--image similarity (\textbf{C-T}) between each frame and its corresponding story description. In the single-character setting, D-I and C-I are computed against the target character references. In the multi-character setting, they are computed against the references of each character appearing in the prompt and then averaged across characters. For each story world, the single-character setting evaluates each character on five stories, each of which contains only that character, whereas the multi-character setting uses five stories involving multiple characters.
 We generate five sequences per story and average all metrics over the resulting frames.

\subsection{Single-Character Comparisons}
\subsubsection{Quantitative Experiments.}
As shown in Tab.~\ref{tab:sin_quantitative}, DisEnvisioner achieves the highest C-I on \textit{Frozen}, but its relatively low C-T indicates weaker adaptation to story semantics. Its lower identity scores on \textit{Pororo} further suggest difficulty in capturing the distinctive attributes of highly stylized characters. In contrast, Textual Inversion achieves the highest C-T, but its substantially lower D-I and C-I reveal limited preservation of fine-grained character appearance. Other tuning-based and CL methods generally show stronger identity preservation. For example, Orth-Adapt provides more balanced performance and is particularly competitive on \textit{Pororo}, but requires post-hoc merging of separately learned character adaptations. EvoTale achieves a stronger overall balance without such model composition, obtaining the highest D-I in SD1.5 comparison. It improves over the strongest competing method by $+4.71\%$ on \textit{Pororo} and $+3.77\%$ on \textit{Frozen}, while maintaining competitive C-I and C-T.

Among the cross-backbone references, CharaConsist shows weaker reference-identity alignment due to its focus on cross-image consistency rather than explicit reference-appearance modeling. FLUX.1 Kontext achieves the highest D-I on \textit{Pororo}. However, as shown in Figs.~\ref{fig:fig1} and~\ref{fig:exp1_single}, this result partly reflects overly literal preservation of the reference image with limited story-conditioned adaptation. Ours maintains strong identity preservation on both datasets with favorable C-T, indicating a better balance between reference fidelity and story adaptation.

\begin{figure*}[!t]
    \centering
    \includegraphics[width=1.0\textwidth]{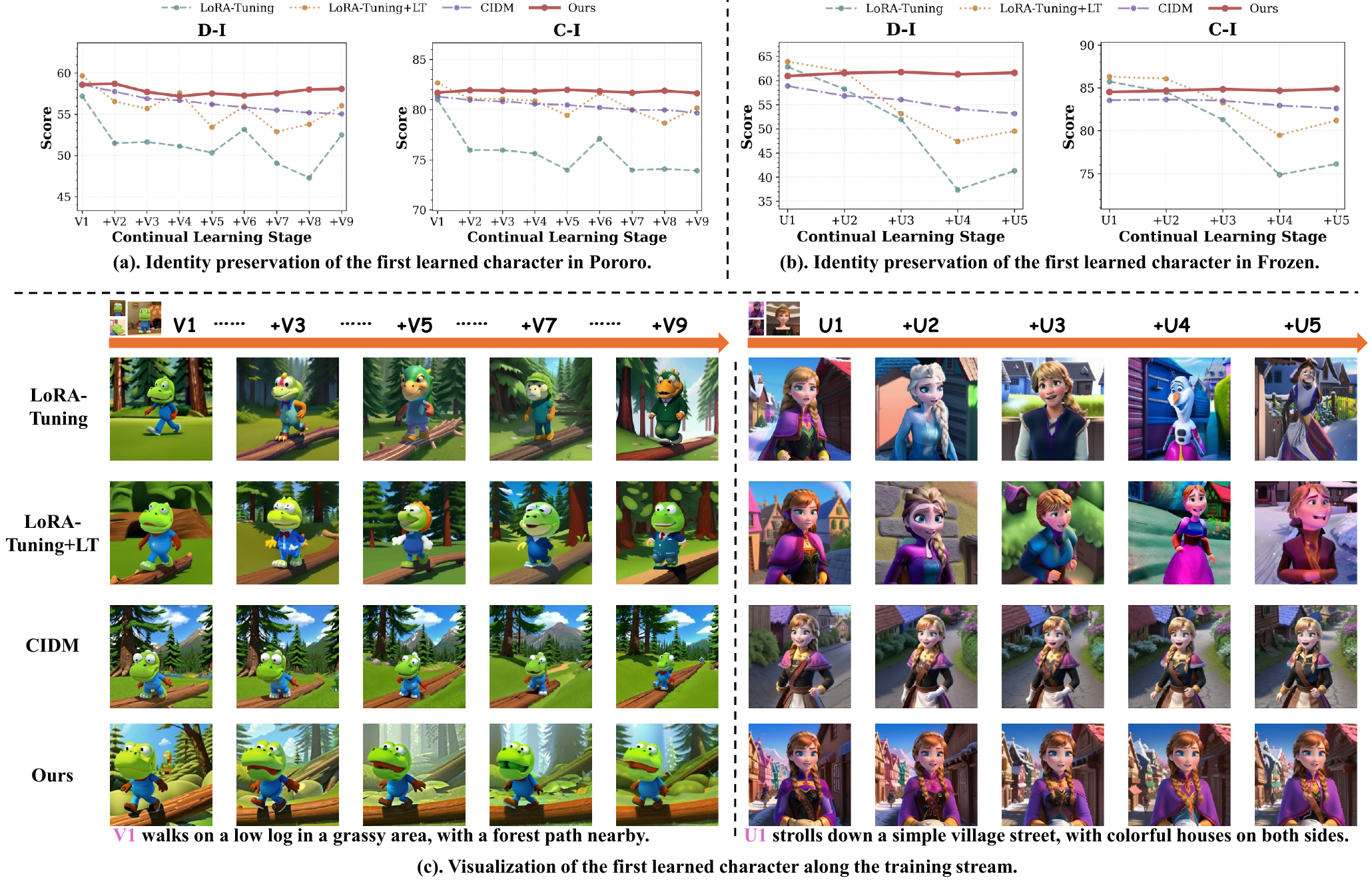}
\caption{
\textbf{Identity preservation during continual integration.}
\textbf{(a)--(b)} Identity scores of the first learned character across learning stages on \textit{Pororo} and \textit{Frozen}.
\textbf{(c)} Qualitative results across stages, with arrows indicating the learning order.
}
    \label{fig:exp2_cl_first}
\end{figure*}

\subsubsection{Qualitative Experiments.}
We additionally include StoryWeaver and DreamStory, whose outputs use different aspect ratios, for qualitative comparisons in Fig.~\ref{fig:exp1_single}. The qualitative results further explain the performance differences observed in Tab.~\ref{tab:sin_quantitative}. DisEnvisioner often produces portrait-like compositions and retains reference-specific poses. For example, its outputs for V6 in \textit{Pororo} and U5 in \textit{Frozen} remain close to the reference viewpoint and pose, despite being conditioned on new story contexts. Such strong adherence to the reference helps preserve character similarity but limits adaptation to diverse story contexts. On \textit{Pororo}, it also captures distinctive character attributes less precisely, consistent with its weaker identity scores. Textual Inversion better follows the requested scenes but often loses fine-grained appearance details. FLUX.1 Kontext also shows strong scene alignment, yet occasionally preserves reference-specific cues too literally. For example, V2 retains the roller skates from the reference despite the absence of any skating-related context in the story prompt, with the main changes concentrated in the surrounding scene. DreamStory and CharaConsist primarily emphasize maintaining an established character appearance across generated images rather than explicitly matching the user-provided reference character, which can lead to identity deviations in some cases. Other tuning-based and continual customization methods generally improve character fidelity but may still exhibit appearance drift across scenes and poses. In comparison, EvoTale more consistently preserves fine-grained character identities while naturally adapting them to diverse story scenes, poses, and actions.

\subsection{Character Identity Preservation}
\label{sec:exp_cl}
To further assess character identity preservation during story-world expansion, we conduct two complementary analyses, \emph{i.e.,} identity retention across sequential updates and stability under multi-character integration.

\subsubsection{Continual Identity Preservation.}
We evaluate accumulated forgetting by tracking the first learned character, which experiences the longest sequence of subsequent updates, as shown in Fig.~\ref{fig:exp2_cl_first}. We compare EvoTale with CIDM and two sequential LoRA baselines: \emph{LoRA-Tuning}, which sequentially updates a shared LoRA module, and \emph{LoRA-Tuning+LT}, which additionally learns a character-specific token embedding. Since these methods differ in their adaptation mechanisms and capacity allocation, we focus on identity stability across learning stages rather than absolute scores at individual stages.
\begin{table*}[!h]
    \centering
    \renewcommand{\arraystretch}{0.97}
\caption{
Before-and-after comparison of character customization performance.
For Mix-of-Show and Orth-Adapt, \emph{Before} denotes independently trained character modules, while \emph{After} denotes the merged model.
For EvoTale, \emph{Before} denotes the average performance of each character immediately after its introduction, while \emph{After} denotes its average performance in the final continually updated model.
$\Delta$ denotes \emph{After} minus \emph{Before}.
}
    \begin{tabular*}{\textwidth}{@{\extracolsep{\fill}}llccccccccc@{}}
    \toprule
    \textbf{Dataset} & \textbf{Method}
    & \multicolumn{3}{c}{D-I$\uparrow$}
    & \multicolumn{3}{c}{C-I$\uparrow$}
    & \multicolumn{3}{c}{C-T$\uparrow$} \\
    \cmidrule(lr){3-5} \cmidrule(lr){6-8} \cmidrule(lr){9-11}
    & & Before & After & $\Delta$
    & Before & After & $\Delta$
    & Before & After & $\Delta$ \\
    \midrule

    \multirow{3}{*}{\textit{Pororo}}
    & Mix-of-Show
    & 53.40 & 49.89 & -3.51
    & 81.13 & 79.38 & -1.75
    & 32.71 & 34.17 & +1.46 \\

    & Orth-Adapt
    & 54.18 & 52.01 & -2.17
    & 80.76 & 79.99 & -0.77
    & 33.60 & 34.06 & +0.46 \\

    & EvoTale
    & 54.85 & 54.46 & -0.39
    & 80.92 & 80.89 & -0.03
    & 34.73 & 34.42 & -0.31 \\

    \midrule

    \multirow{3}{*}{\textit{Frozen}}
    & Mix-of-Show
    & 49.99 & 45.05 & -4.94
    & 83.26 & 80.66 & -2.60
    & 34.46 & 35.08 & +0.62 \\

    & Orth-Adapt
    & 51.96 & 49.61 & -2.35
    & 84.51 & 82.56 & -1.95
    & 34.65 & 35.10 & +0.45 \\

    & EvoTale
    & 52.31 & 51.50 & -0.81
    & 83.12 & 82.84 & -0.28
    & 35.42 & 35.23 & -0.19 \\

    \bottomrule
    \end{tabular*}
    \label{tab:before_after_merge}
\end{table*}
As shown in Fig.~\ref{fig:exp2_cl_first}(a) and Fig.~\ref{fig:exp2_cl_first}(b), across both \textit{Pororo} and \textit{Frozen}, the two sequential LoRA baselines exhibit noticeable fluctuations and an overall decline in D-I and C-I as new characters are introduced. This behavior indicates accumulated interference from subsequent updates. CIDM is comparatively stable, but its identity scores still gradually decrease along the learning stream. In contrast, EvoTale maintains stable D-I and C-I scores with only minor variations across stages. The qualitative comparisons in Fig.~\ref{fig:exp2_cl_first}(c) further show that sequential LoRA updates may alter defining character attributes or cause identity drift toward newly learned characters, whereas EvoTale more consistently preserves the first character throughout the continual learning stream.

\begin{figure*}[!t]
    \centering
    \includegraphics[width=1.0\textwidth]{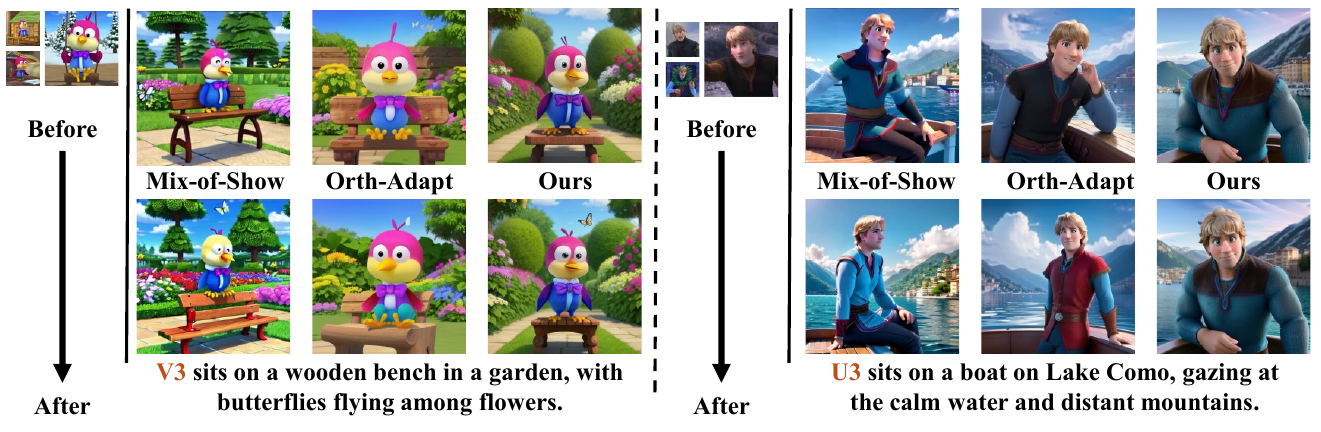}
    \caption{
    \textbf{Character identity before and after multi-character integration.}
    EvoTale exhibits smaller identity changes than post-hoc composition methods after integrating multiple characters.
    }
    \label{fig:exp2_cl_merge}
\end{figure*}

\subsubsection{Character Customization Stability.}
We further examine whether learned character identities remain stable during multi-character integration. As shown in Tab.~\ref{tab:before_after_merge}, both composition-based methods exhibit identity degradation after model composition, with Orth-Adapt showing better stability. In comparison, EvoTale shows substantially smaller before-to-after changes across both story worlds. The qualitative examples in Fig.~\ref{fig:exp2_cl_merge} show a similar trend. For example, Mix-of-Show changes V3's head coloration and U3's facial appearance, while Orth-Adapt changes V3's clothing. In contrast, EvoTale better preserves character identity as more characters are integrated. These results indicate that organizing character-specific knowledge within a unified LoRA branch through orthogonal-basis-guided routing enables more stable story-world expansion without post-hoc module composition.

\begin{table*}[!t]
    \centering
    \small
    \setlength{\tabcolsep}{4pt}
    \renewcommand{\arraystretch}{1.08}

    \begin{threeparttable}

\caption{
Quantitative comparison on multi-character story visualization.
The upper block compares methods in SD1.5, while the lower block reports SDXL reference results.
Best results are shown in \textbf{bold} and the second-best are \underline{underlined}, while only the best result is highlighted in the SDXL block.
$\dagger$ denotes SDXL-based models; unmarked methods use SD1.5.
}
    \label{tab:multi_quantitative}
    \begin{tabular*}{\textwidth}{
        @{\extracolsep{\fill}}
        llcccccc
        @{}
    }
    \toprule
        \textbf{Type}
        & \textbf{Method}
        & \multicolumn{3}{c}{\textbf{Pororo}}
        & \multicolumn{3}{c}{\textbf{Frozen}}
        \\
    \cmidrule(lr){3-5}
    \cmidrule(lr){6-8}
        &
        & D-I$\uparrow$
        & C-I$\uparrow$
        & C-T$\uparrow$
        & D-I$\uparrow$
        & C-I$\uparrow$
        & C-T$\uparrow$
        \\

    \midrule
        \multicolumn{8}{c}{\textit{SD1.5-Based Comparison}} \\
    \midrule

        \multirow{5}{*}{Tuning-based}
        & LoRA-S
        & 37.24
        & 65.49
        & 33.60
        & 37.47
        & 72.13
        & 33.90
        \\

        & LoRA-C
        & 37.31
        & 65.54
        & 33.85
        & 42.38
        & 74.20
        & 34.28
        \\

        & Mix-of-Show
        & 40.27
        & 69.13
        & 34.72
        & \underline{43.57}
        & 74.34
        & 36.11
        \\

        & Mix-of-Show$^{*}$
        & 36.23
        & 68.04
        & 34.56
        & 39.24
        & 75.49
        & 36.08
        \\

        & LoRA-Composer
        & 40.00
        & 67.69
        & \underline{36.56}
        & 41.35
        & \underline{76.18}
        & \textbf{37.17}
        \\

        Story visualization(Tuning-based) & StoryWeaver
        & \textbf{53.38}
        & \textbf{72.61}
        & 34.03
        & 42.19
        & 75.98
        & 36.15
        \\

        CL
        & Ours
        & \underline{44.95}
        & \underline{69.54}
        & \textbf{36.59}
        & \textbf{45.32}
        & \textbf{77.98}
        & \underline{36.44}
        \\

    \midrule
        \multicolumn{8}{c}{\textit{SDXL Reference}} \\
    \midrule

        Story visualization(Training-free)
        & DreamStory$^\dagger$
        & 31.50
        & 63.39
        & \textbf{37.14}
        & 39.68
        & 73.11
        & 36.55
        \\

        CL
        & Ours$^\dagger$
        & \textbf{46.45}
        & \textbf{71.04}
        & 36.66
        & \textbf{46.35}
        & \textbf{79.20}
        & \textbf{36.78}
        \\

    \bottomrule
    \end{tabular*}

    \begin{tablenotes}[flushleft]
        \footnotesize
        \item[$*$] For Mix-of-Show$^{*}$, the pose-conditioning weight is set to zero while retaining the key-pose adapter computation.
    \end{tablenotes}

    \end{threeparttable}
\end{table*}
\begin{figure*}[!t]
    \centering
    \includegraphics[width=1.0\textwidth]{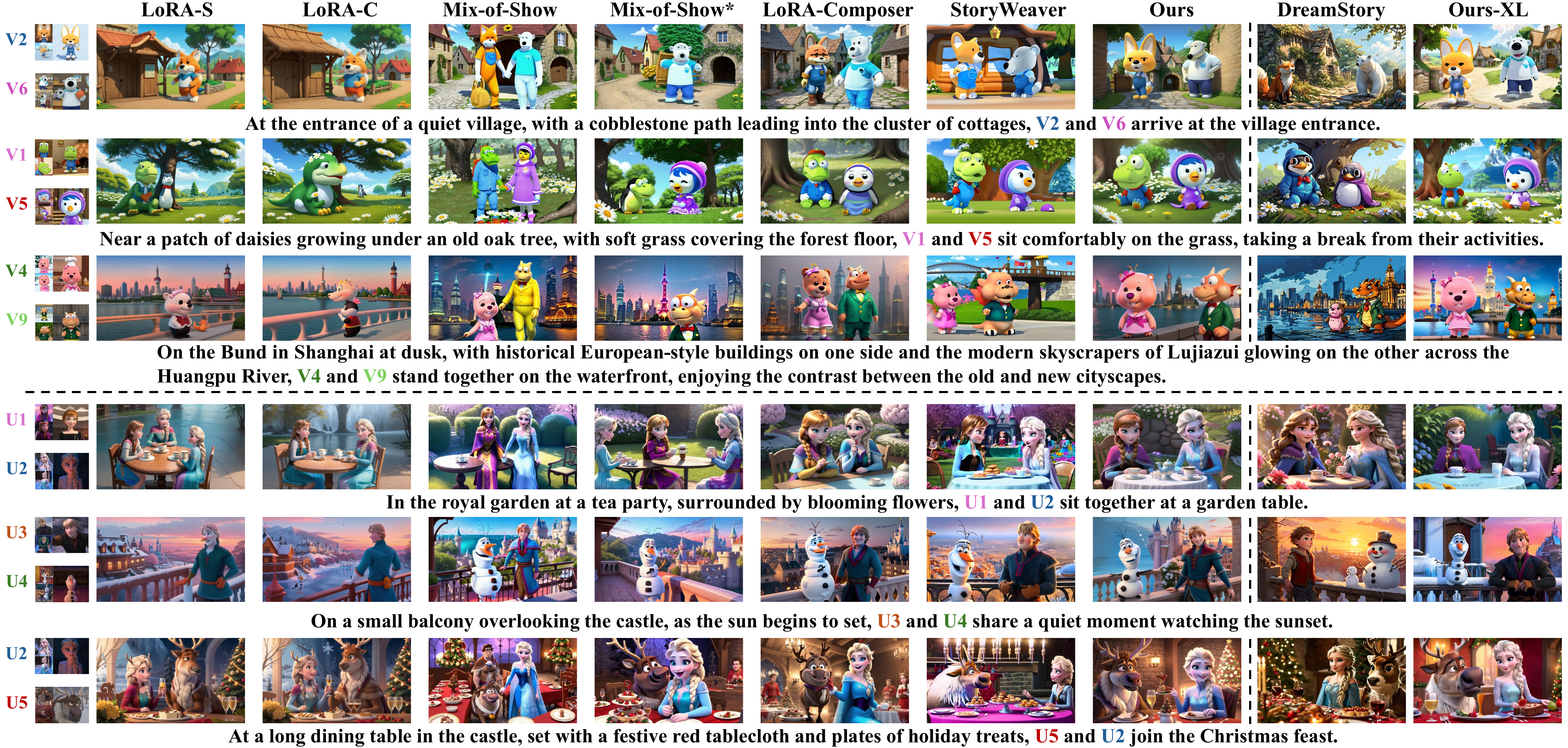}
    \caption{
    Qualitative comparison of multi-character story visualization.
    EvoTale better preserves distinct character identities while maintaining natural interactions and strong alignment with the story descriptions.
    }
    \label{fig:exp3_multi}
\end{figure*}
\subsection{Multi-Character Comparisons}

\subsubsection{Quantitative Experiments.}
Quantitative results for multi-character visual storytelling are shown in Tab.~\ref{tab:multi_quantitative}. LoRA-S alternates among character-specific LoRAs during denoising, whereas LoRA-C aggregates their predictions at each step. Without explicit character-specific spatial routing, both methods remain susceptible to identity confusion in multi-character scenes, as reflected by their relatively low D-I scores on \textit{Pororo}. Mix-of-Show combines customized concepts through post-hoc gradient fusion and spatial constraints. When pose conditioning is disabled in Mix-of-Show*, its lower D-I indicates weaker identity preservation. Additionally, merging five characters also takes about 11 minutes in our setting. LoRA-Composer assigns individual LoRAs to designated regions, but its latent re-initialization and constraint-guided refinement increase generation time to about 60\,s per image. StoryWeaver preserves identity through knowledge-enhanced spatial guidance to inject character semantics, yet its implicit spatial guidance remains limited in balancing character fidelity and new story semantics.

In comparison, EvoTale achieves a better balance between character identity and story-text alignment. Under SD1.5, it maintains competitive identity preservation with strong story-text alignment on \textit{Pororo}, and particularly strong identity preservation on \textit{Frozen} without sacrificing semantic alignment. It requires about 30\,s per image without post-hoc model fusion or iterative latent refinement. Under SDXL, DreamStory shows favorable story-text alignment but weaker reference-identity preservation and requires about 88\,s per image, whereas EvoTale-SDXL takes about 68\,s while achieving substantially higher D-I and C-I with comparable semantic alignment. All inference times are measured on a single GeForce RTX 4090 GPU.

\subsubsection{Qualitative Experiments.}
As shown in Fig.~\ref{fig:exp3_multi}, existing methods exhibit different limitations in multi-character story visualization. LoRA-S and LoRA-C tend to omit required characters or to exhibit identity blending among characters within the same frame. The key-pose guidance in Mix-of-Show helps maintain character placement but can result in more constrained compositions. Without pose guidance, Mix-of-Show$^{*}$ allows greater layout flexibility but becomes more susceptible to character omission or identity deviation. LoRA-Composer and StoryWeaver generally render both characters within coherent scenes, although identity deviations remain in some cases. For example, LoRA-Composer loses some distinctive appearance details of V5 in the $2^{nd}$ example, while the visual traits of U1 and U2 become partially entangled in StoryWeaver's tea-party result in the $4^{th}$ row. DreamStory follows the requested scenes and character interactions well, but tends to generate semantically plausible characters rather than faithfully preserving the specified identities. In comparison, our method more consistently retains the distinct identities of the prompted characters while depicting the requested scenes and interactions. Overall, EvoTale achieves a favorable balance between multi-character identity fidelity and story-text alignment.
\begin{table}[t]
    \centering
    \scriptsize
    \setlength{\tabcolsep}{3pt}
    \renewcommand{\arraystretch}{1.10}
    
\caption{
User study on single- and multi-character story visualization.
C-A, S-A, and V-Q denote character alignment, story alignment, and visual quality, respectively.
Best results are shown in \textbf{bold} and the second-best are \underline{underlined}.
$\star$ denotes DiT-based models and $\dagger$ denotes SDXL-based models;
unmarked methods use SD1.5.
}

    \begin{tabular*}{\columnwidth}{
        @{\extracolsep{\fill}}
        lccc
        @{}
    }
    \toprule
    \textbf{Method}
    & \textbf{C-A}$\uparrow$
    & \textbf{S-A}$\uparrow$
    & \textbf{V-Q}$\uparrow$ \\
    \midrule

    \multicolumn{4}{c}{\textit{Single Character}} \\
    \midrule

    DreamStory$^\dagger$
    & 1.56 & 3.52 & 3.80 \\

    CharaConsist$^\star$
    & 1.80 & 3.44 & 3.56 \\

    DisEnvisioner
    & 1.44 & 2.36 & 2.12 \\

    FLUX.1 Kontext$^\star$
    & 3.12 & \underline{3.76} & 3.28 \\

    Mix-of-Show
    & 2.64 & 2.88 & 2.92 \\

    Orth-Adapt
    & 3.20 & 3.04 & 3.12 \\

    StoryWeaver
    & 3.96 & 3.44 & 3.60 \\

    Textual Inversion
    & 2.44 & 2.88 & 2.32 \\

    CIDM
    & 2.48 & 2.84 & 2.60 \\

    Ours
    & \underline{4.24}
    & 3.68
    & \underline{3.84} \\

    Ours-XL$^\dagger$
    & \textbf{4.48}
    & \textbf{3.84}
    & \textbf{4.08} \\

    \midrule
    \multicolumn{4}{c}{\textit{Multi-Character}} \\
    \midrule

    DreamStory$^\dagger$
    & 1.52 & \underline{4.12} & 3.68 \\

    LoRA-S
    & 1.32 & 1.24 & 2.72 \\

    LoRA-C
    & 1.40 & 1.84 & 2.80 \\

    Mix-of-Show
    & 2.44 & 2.92 & 2.76 \\

    Mix-of-Show*
    & 2.32 & 2.68 & 2.60 \\

    LoRA-Composer
    & 2.72 & 2.92 & 3.16 \\

    StoryWeaver
    & 3.72 & 3.64 & 3.56 \\

    Ours
    & \underline{4.08}
    & 3.92
    & \textbf{4.20} \\

    Ours-XL$^\dagger$
    & \textbf{4.28}
    & \textbf{4.16}
    & \underline{4.12} \\

    \bottomrule
    \end{tabular*}
    \label{tab:user_study}
\end{table}

\begin{table*}[!t]
    \centering
    \small
    \setlength{\tabcolsep}{4pt}
    \renewcommand{\arraystretch}{1.08}
\caption{
Ablation study of the main components and CI parameterization on single-character story visualization.
For each dataset, best results are shown in \textbf{bold} and the second-best are \underline{underlined}.
LT denotes learnable token embeddings.
}
    \begin{tabular*}{\textwidth}{@{\extracolsep{\fill}}lcccccc@{}}
    \toprule
    \textbf{Setting}
    & \multicolumn{3}{c}{\textbf{Pororo}}
    & \multicolumn{3}{c}{\textbf{Frozen}} \\
    \cmidrule(lr){2-4}
    \cmidrule(lr){5-7}
    & D-I$\uparrow$ & C-I$\uparrow$ & C-T$\uparrow$
    & D-I$\uparrow$ & C-I$\uparrow$ & C-T$\uparrow$ \\
    
    \midrule
    \multicolumn{7}{c}{\textit{Effect of Main Components}} \\
    \midrule
    
    Shared LoRA (Sequential, No Replay)
    & 45.75
    & 77.04
    & \textbf{35.61}
    & 42.84
    & 78.75
    & \textbf{36.08} \\
    
    w/o CQG (Fixed Base Budget)
    & 51.85
    & 79.53
    & \underline{35.45}
    & 47.63
    & 80.88
    & \underline{35.73} \\
    
    w/o CQG (Fixed Large Budget)
    & \textbf{59.32}
    & \textbf{83.84}
    & 31.96
    & \textbf{56.42}
    & \textbf{85.03}
    & 33.19 \\
    
    Full Model
    & \underline{54.46}
    & \underline{80.89}
    & 34.42
    & \underline{51.50}
    & \underline{82.84}
    & 35.23 \\
    
    \midrule
    \multicolumn{7}{c}{\textit{Effect of CI Parameterization}} \\
    \midrule
    
    LoRA (Fixed Base Budget)
    & \underline{47.28}
    & \underline{77.06}
    & \underline{35.60}
    & \underline{47.02}
    & \textbf{81.62}
    & 35.53 \\
    
    CI w/o LT (Fixed Base Budget)
    & 46.88
    & 74.78
    & \textbf{35.91}
    & 46.28
    & 79.84
    & \textbf{36.11} \\
    
    CI w/ LT (Fixed Base Budget)
    & \textbf{51.85}
    & \textbf{79.53}
    & 35.45
    & \textbf{47.63}
    & \underline{80.88}
    & \underline{35.73} \\
    
    \bottomrule
    \end{tabular*}
    \label{tab:ablation_single}
\end{table*}
\begin{table*}[!t]
    \centering
    \small
    \setlength{\tabcolsep}{4pt}
    \renewcommand{\arraystretch}{1.08}
\caption{
Ablation study of CA-RF on multi-character story visualization.
Best results are shown in \textbf{bold} and the second-best are \underline{underlined}.
RS-CD and IA-GD denote Role-Specific Crop Denoising and Identity-Aware Global Denoising, and
Global denotes standard global denoising.
Cross-AR and Regional-SA denote character-specific cross-attention routing and region-restricted self-attention, respectively.
}
    
    \begin{tabular*}{\textwidth}{@{\extracolsep{\fill}}lcccccc@{}}
    \toprule
    \textbf{Setting}
    & \multicolumn{3}{c}{\textbf{Pororo}}
    & \multicolumn{3}{c}{\textbf{Frozen}} \\
    \cmidrule(lr){2-4}
    \cmidrule(lr){5-7}
    & D-I$\uparrow$ & C-I$\uparrow$ & C-T$\uparrow$
    & D-I$\uparrow$ & C-I$\uparrow$ & C-T$\uparrow$ \\
    \midrule

    RS-CD + Global (Fixed Fusion)
    & 42.87
    & 68.18
    & 36.43
    & 43.32
    & 76.19
    & 36.05 \\

    RS-CD + Global (Time-Aware Fusion)
    & 37.29
    & 65.95
    & 36.31
    & 38.85
    & 75.11
    & 35.54 \\

    RS-CD + IA-GD (Cross-AR)
    & 42.68
    & 68.13
    & \textbf{36.79}
    & 42.53
    & \underline{77.16}
    & 36.04 \\

    RS-CD + IA-GD (Cross-AR + Regional-SA)
    & \textbf{46.35}
    & \textbf{69.73}
    & 36.25
    & \textbf{45.65}
    & 76.97
    & \textbf{36.57} \\

    Full CA-RF
    & \underline{44.95}
    & \underline{69.54}
    & \underline{36.59}
    & \underline{45.32}
    & \textbf{77.98}
    & \underline{36.44} \\

    \bottomrule
    \end{tabular*}
    \label{tab:ablation_multi}
\end{table*}
\subsection{User Study}
We further conduct a user study with 25 participants to assess the perceptual quality of different story visualization methods. Participants assign ratings for each method based on anonymously presented generated sequences, considering Character Alignment (C-A), Story Alignment (S-A), and Overall Visual Quality (V-Q) on a five-point Likert scale, with higher scores indicating better performance. As shown in Tab.~\ref{tab:user_study}, EvoTale receives consistently high ratings across the three criteria, indicating favorable human evaluations of EvoTale in terms of character alignment, story alignment, and overall visual quality.
\begin{figure*}[!t]
    \centering
    \includegraphics[width=\textwidth]{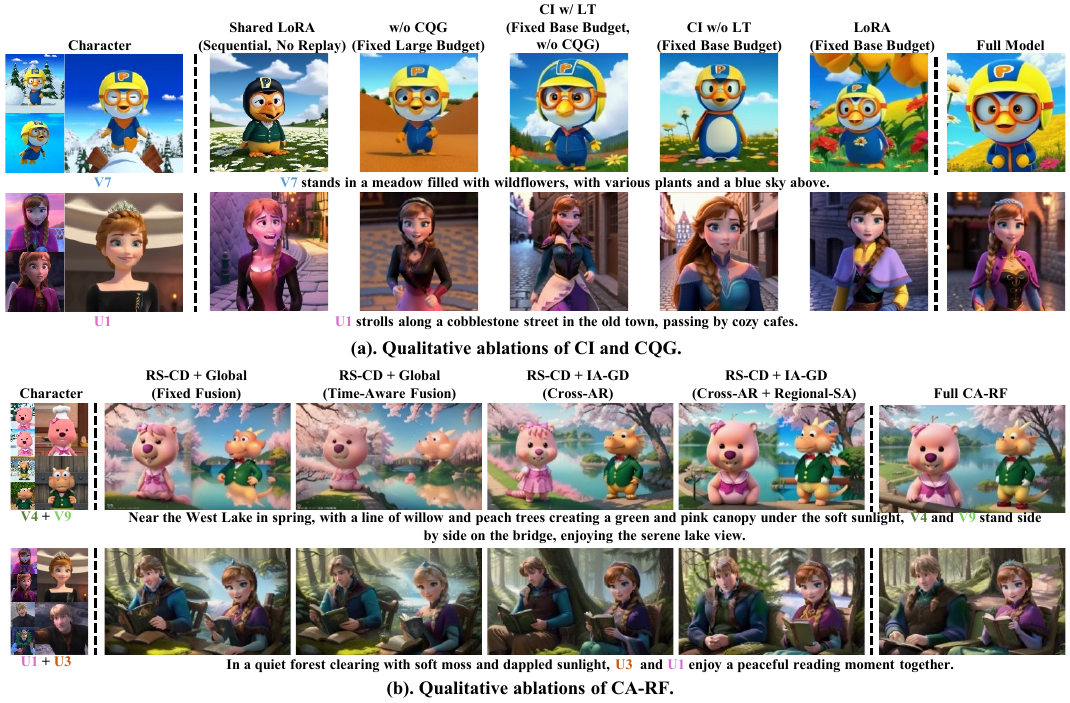}
\caption{
Qualitative ablation results for the key components of EvoTale.
\textbf{(a)} CI and CQG in single-character story settings.
\textbf{(b)} CA-RF sampling in multi-character story settings.
}
    \label{fig:abla}
\end{figure*}

\subsection{Ablation Study}
\subsubsection{Quantitative Analysis}
\textbf{Single-character setting.}
We first evaluate CI and CQG in Tab.~\ref{tab:ablation_single}. To ablate CI, we use a single shared LoRA branch that is sequentially updated for all characters without replaying data from previously learned characters. All other training settings are kept unchanged. This baseline decreases D-I and C-I on both \textit{Pororo} and \textit{Frozen}, demonstrating that CI's character-specific residual accumulation and routing are important for limiting cross-character interference and preserving previously learned identities.
To evaluate CQG, we compare it with two fixed schedules representing relatively conservative and intensive customization budgets. Specifically, each character-specific training set is repeated 500 and 1,500 times under the base and large budgets, respectively. Without CQG, the fixed base budget yields lower identity scores than the full model, indicating insufficient adaptation for some characters. Increasing the fixed budget improves D-I and C-I but noticeably degrades C-T, suggesting that more intensive optimization increasingly favors reference appearance over story-text alignment. By adjusting the refinement budget according to the assessed customization quality, CQG achieves a more favorable balance between identity preservation and semantic alignment.

The lower block examines the parameterization of CI. For a fair comparison, standard LoRA and all CI variants are applied to the same layers and trained with the same base budget. With the shared factor $B$ fixed, CI without learnable tokens obtains lower D-I and C-I than standard LoRA on both story worlds, indicating reduced flexibility in character-specific adaptation. Character-specific learnable token embeddings compensate for this constraint by providing an additional pathway for identity representation. With these embeddings, CI achieves higher D-I than standard LoRA on both story worlds and higher C-I on \textit{Pororo}, while remaining close on \textit{Frozen}.

\noindent \textbf{Multi-character setting.}
Tab.~\ref{tab:ablation_multi} evaluates the key components of Character-Aware Region-Focus Sampling. Compared with fixed fusion, time-aware fusion between RS-CD and standard global denoising decreases identity preservation. Replacing the standard global branch with IA-GD using Cross-AR recovers both identity and semantic alignment, while further restricting self-attention to character regions improves D-I on both story worlds. Full CA-RF introduces a local-to-global self-attention transition, allowing character features to interact with the surrounding scene as denoising proceeds. Although keeping self-attention region-restricted throughout denoising yields slightly higher scores on several metrics, the qualitative results in Fig.\ref{fig:abla}(b) show that it is more susceptible to rigid character layouts. Considering both quantitative performance and scene composition, we adopt full CA-RF as the final sampling strategy.

\subsubsection{Qualitative Analysis}
As shown in Fig.~\ref{fig:abla}(a), using a single shared LoRA branch causes visible identity drift and loss of fine-grained attributes. Without CQG, the fixed base budget may provide insufficient adaptation for some characters, whereas the fixed large budget better reproduces character appearance but can weaken scene details and semantic alignment. For example, the fixed large budget preserves the character’s appearance well but fails to generate ``meadow filled with wildflowers'' described in the  story prompt for V7. Regarding CI parameterization, fixing the shared factor $B$ without learnable tokens limits character-specific adaptation and leads to greater loss of reference-specific identity details than standard LoRA. Introducing character-specific learnable tokens into CI compensates for this constraint and recovers finer identity cues. In contrast, the full setting better balances identity preservation and story alignment, showing the complementary effects of character-specific subspace assignment, learnable tokens, and adaptive budget allocation.

Results in Fig.~\ref{fig:abla}(b) further illustrate the effects of different CA-RF components. Fixed fusion between RS-CD and standard global denoising provides strong regional identity guidance but tends to produce relatively rigid character layouts. Introducing time-aware fusion increases the contribution of global denoising and improves composition flexibility, but identity preservation becomes weaker when the global branch lacks identity-aware routing. Replacing standard global denoising with IA-GD using Cross-AR improves character assignment and global scene coordination. Further restricting self-attention within character regions strengthens identity preservation and produces more natural compositions than fixed fusion. However, the persistent regional separation can introduce visible seam-like boundaries between character regions, as observed in both examples in Fig.~\ref{fig:abla}(b). By allowing character features to progressively interact with the surrounding scene during denoising, Full CA-RF reduces such region-wise discontinuities and yields more spatially coherent compositions while preserving the prompted identities. These observations are consistent with the quantitative results in Tab.~\ref{tab:ablation_multi}.

%% file: sec/5_con.tex
\section{Conclusion}
In this paper, we present EvoTale, a continual character customization framework for expanding stylized story worlds. We introduce the All-in-One-World Character Integrator to accumulate character-specific residual components within a unified LoRA branch through orthogonal-basis-guided routing, while freezing previously learned components to limit cross-character interference. We further develop the Character Quality Gate to adaptively control the optimization budget through rubric-guided MLLM assessment, and Character-Aware Region-Focus Sampling to combine regional identity guidance with identity-aware global denoising for coherent multi-character generation. Experimental results demonstrate a favorable balance across character fidelity, story-text alignment, continual identity retention, and multi-character generation quality.